%% file: PromptSR.tex
\documentclass[10pt,twocolumn,letterpaper]{article}

\usepackage[pagenumbers]{iccv} 


\usepackage{graphicx}
\usepackage{booktabs}
\usepackage{amsmath}
\usepackage{amssymb}
\usepackage{multirow}
\usepackage{adjustbox}  
\usepackage{times,tabulary,overpic}
\usepackage{colortbl}
\usepackage{makecell}
\usepackage{float}
\usepackage{url}
\usepackage{balance}
\usepackage{bbding}
\usepackage[utf8]{inputenc} 
\usepackage[T1]{fontenc}    
\usepackage{wrapfig}
\usepackage{subcaption}

\definecolor{color3}{rgb}{0.95,0.95,0.95}
\definecolor{iccvblue}{rgb}{0.21,0.49,0.74}
\usepackage[pagebackref,breaklinks,colorlinks,allcolors=iccvblue]{hyperref}

\def\paperID{***} 
\def\confName{ICCV}
\def\confYear{2025}

\title{Image Super-Resolution with Text Prompt Diffusion}

\author{
Zheng Chen$^{1}$,\enspace 
Yulun Zhang$^{1}$\thanks{Corresponding authors: Yulun Zhang, Linghe Kong}~,\enspace 
Jinjin Gu$^{2,3}$,\enspace 
Xin Yuan$^{4}$,\enspace \\ 
Linghe Kong$^{1}$\footnotemark[1]~,\enspace 
Guihai Chen$^{1}$,\enspace 
Xiaokang Yang$^{1}$ \\
\textsuperscript{1}Shanghai Jiao Tong University,\enspace 
\textsuperscript{2}Shanghai AI Laboratory,\enspace \\ 
\textsuperscript{3}The University of Sydney,\enspace 
\textsuperscript{4}Westlake University
\vspace{-4.mm}
}

\begin{document}

\maketitle

\begin{abstract}
Image super-resolution (SR) methods typically model degradation to improve reconstruction accuracy in complex and unknown degradation scenarios. However, extracting degradation information from low-resolution images is challenging, which limits the model performance. To boost image SR performance, one feasible approach is to introduce additional priors. Inspired by advancements in multi-modal methods and text prompt image processing, we introduce text prompts to image SR to provide degradation priors. Specifically, we first design a text-image generation pipeline to integrate text into the SR dataset through the text degradation representation and degradation model. By adopting a discrete design, the text representation is flexible and user-friendly. Meanwhile, we propose the PromptSR to realize the text prompt SR. The PromptSR leverages the latest multi-modal large language model (MLLM) to generate prompts from low-resolution images. It also utilizes the pre-trained language model (e.g., T5 or CLIP) to enhance text comprehension. We train the PromptSR on the text-image dataset. Extensive experiments indicate that introducing text prompts into SR, yields impressive results on both synthetic and real-world images. Code:~\url{https://github.com/zhengchen1999/PromptSR}.
\end{abstract}

\setlength{\abovedisplayskip}{1pt}
\setlength{\belowdisplayskip}{1pt}

\vspace{-6.mm}
\section{Introduction}
\vspace{-2.mm}
Single image super-resolution (SR) aims to recover high-resolution (HR) images from their corresponding low-resolution (LR) counterparts. Over recent years, the proliferation of deep learning-based methods~\cite{dong2014learning,zhang2018image,chen2023dual} has significantly advanced this domain.
Nevertheless, the majority of these methods are trained with known degradation (\eg, bicubic interpolation), which limits their generalization capabilities~\cite{wang2021real,zhang2023crafting}. 
Consequently, these methods face challenges when applied to scenarios with complex and diverse degradations, such as real-world applications.

A feasible approach to tackle the diverse SR challenges is blind SR. Blind SR focuses on reconstructing LR images with complex and unknown degradation, making it suitable for various scenarios~\cite{liu2022blind}.
Methods within this realm can roughly be divided into several categories.
\textbf{(1)} Explicit methods~\cite{zhang2018learning} typically rely on predefined degradation models. They estimate degradation parameters (\eg, blur kernel or noise) as conditional inputs to the SR model. However, the predefined degradation models exhibit a limited degradation representation scope, restricting the generality of methods.
\textbf{(2)} Implicit methods~\cite{cai2019toward,wei2021unsupervised} capture underlying degradation models through extensive external datasets. They achieve this by leveraging real-captured HR-LR image pairs, or HR and unpaired LR data, to learn the data distribution. Nevertheless, learning the data distribution is challenging, with unsatisfactory results.
\textbf{(3)} Currently, another image SR paradigm~\cite{zhang2021designing,wang2021real} is popularized: defining complex degradation to synthesize a large amount of data for training. To simulate real-world degradation, these approaches set the degradation distribution sufficiently extensive. Nonetheless, this increases the learning difficulty of the SR model and causes a performance drop.

\begin{figure}[t]
\scriptsize
\centering
\begin{tabular}{cc}

\hspace{-4.5mm}
\begin{adjustbox}{valign=t}
\begin{tabular}{c}
\includegraphics[width=0.135\textwidth]{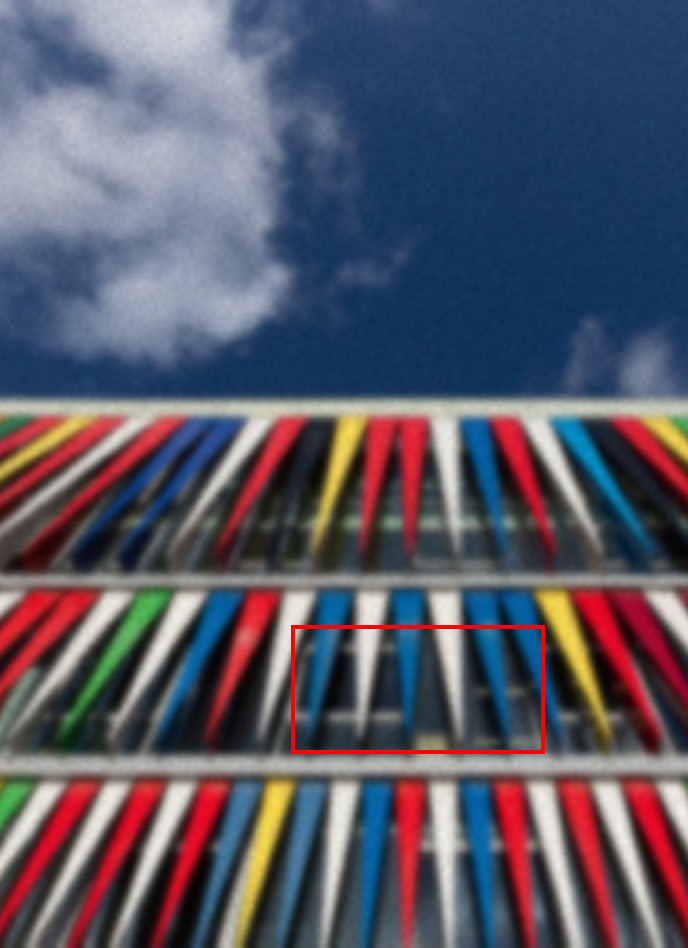}
\\
LR
\end{tabular}
\end{adjustbox}

\hspace{-0.46cm}
\begin{adjustbox}{valign=t}
\begin{tabular}{cccc}
\includegraphics[width=0.165\textwidth]{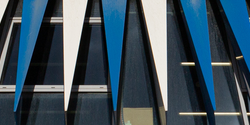} \hspace{-4mm} &
\includegraphics[width=0.165\textwidth]{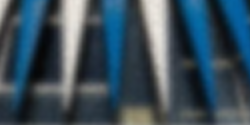} \hspace{-4mm} &
\\
HR  \hspace{-4mm} &
Bicubic \hspace{-4mm} &
\\
\includegraphics[width=0.165\textwidth]{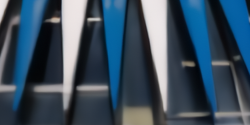} \hspace{-4mm} &
\includegraphics[width=0.165\textwidth]{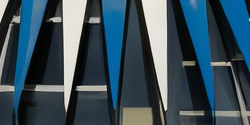} \hspace{-4mm} &
\\ 
w/o text prompt  \hspace{-4mm} &
w/ text prompt\hspace{-4mm}
\\
\end{tabular}
\end{adjustbox}

\end{tabular}
\vspace{-3mm}
\caption{Visual comparison ($\times$4). The LR image undergoes complex and unknown degradations (\eg, blur, noise, and downsampling). By introducing text prompts (\eg, [\textbf{\textit{heavy blur, upsample, medium noise, medium compression, downsample}}], in the instance) into the SR task to provide degradation priors, the reconstruction quality can be effectively improved.} 
\label{fig:visual-0}
\vspace{-6mm}
\end{figure}

In summary, the modeling of degradation is crucial to image SR, typically in complex application scenarios. However, most methods extract degradation information mainly from LR images, which is challenging and limits performance. One approach to advance SR performance is to introduce additional priors, such as reference priors~\cite{jiang2021robust} or generative priors~\cite{chan2021glean,yang2021gan}. \textbf{Motivated} by recent advancements in the multi-modal model~\cite{radford2021learning,liu2023llava}, text prompt image generation~\cite{ramesh2021zero,rombach2022high}, and manipulation~\cite{brooks2023instructpix2pix}, we introduce the text prompt to provide priors for image SR.
This approach offers several advantages:
\textbf{(1)} Textual information is inherently flexible and suitable for various situations.
\textbf{(2)} The power of the current pre-trained language model can be leveraged.
\textbf{(3)} Text guidance can serve as a complement to current methods for image SR.

In this work, we propose a method to introduce text as additional priors to enhance image SR. 
Our design encompasses two aspects:
\textbf{(1) Dataset:} For text prompt SR, large-scale multi-modal (text-image) data is crucial, yet challenging to collect manually. As mentioned above, the degradation models~\cite{wang2021real} can synthesize vast amounts of HR-LR image pairs. Hence, we incorporate text into the degradation model to generate the corresponding data. 
\textbf{(2) Model:} Text prompt SR inherently involves text processing. Meanwhile, the pre-trained language models possess powerful textual understanding capabilities. Thus, we utilize these models within our model to enhance text guidance.

Specifically, we develop a text-image generation pipeline that integrates text into the SR degradation model.
\textbf{Text prompt for degradation:} We utilize text prompts to represent the degradation to provide additional prior. Since the LR image could provide the majority of low-frequency~\cite{zhang2018image} and semantic information related to the content~\cite{rombach2022high}, we care little about the abstract description of the overall image. 
\textbf{Text representation:} We first discretize degradation into components (\eg, blur, noise). Then, we employ the binning method~\cite{zhang2023crafting} to partition the degradation distribution, describe each segment textually, and merge them, to get the final text prompt. This discrete approach simplifies representation, which is intuitive and user-friendly to apply.
\textbf{Flexible format:} To enhance prompt practicality, we adopt a more flexible format, such as arbitrary order or simplified (\eg, only noise description) prompts. The recovery results, benefiting from the generalization of prompts, are also remarkable.  
Details are shown in Sec.~\ref{sec:ablation}.
\textbf{Text-image dataset:} We adopt degradation models akin to previous methods~\cite{zhang2021designing,wang2021real} to generate HR-LR image pairs. Simultaneously, we utilize the degradation description approach to produce the text prompts, generating the dataset.

We further propose a network, PromptSR, to implement text prompt image SR. Our PromptSR leverages the diffusion models~\cite{ho2020denoising,saharia2022image} for high-quality image restoration. Concurrently, we employ the latest multi-modal large language model (MLLM)~\cite{liu2023llava,openai2023gpt4,ye2024mplug,wu2024q,you2024descriptive} to generate the corresponding prompt from the LR image. Moreover, as analyzed previously, we apply the pre-trained language model (\eg, T5~\cite{raffel2020exploring} or CLIP~\cite{radford2021learning}) to encode the text prompts. By leveraging the powerful understanding capabilities of language models, PromptSR can model image degradation better, leading to improved results.
We train the PromptSR on the generated text-image dataset, and it performs excellently on both synthetic and real-world images. As illustrated in Fig.~\ref{fig:visual-0}, the model reconstructs a more realistic and clear image when applying the text prompt.

\vspace{0.2em} 
Overall, we summarize the main contributions as:

\begin{itemize}
\vspace{0.2em} 
\item We introduce text prompts as degradation priors to advance image SR. This explores the application of textual information in the image SR task.

\vspace{0.2em} 
\item We develop a text-image generation pipeline that integrates the user-friendly and flexible prompt into the SR dataset via text representation and degradation model.

\vspace{0.2em} 
\item We propose a network, PromptSR, to realize the text prompt SR. The PromptSR utilizes the pre-trained language model to improve the restoration.

\vspace{0.2em} 
\item Extensive experiments show that the introduction of text prompts into image SR leads to impressive results on both synthetic and real-world images.
\end{itemize}

\begin{figure*}[t]
    \centering
    \begin{tabular}{c}
    \hspace{-3.mm}\includegraphics[width=1.\linewidth]{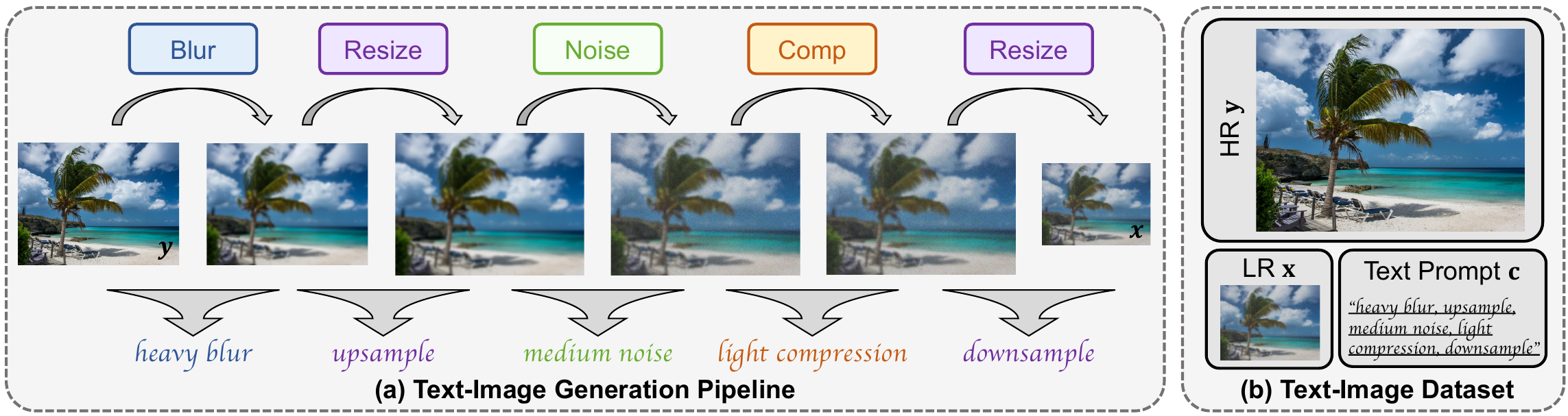} \\
    \end{tabular}
    \vspace{-3.mm}
    \caption{Illustration of the text-image generation pipeline. (a) The pipeline comprises the degradation model (top) and the text representation (bottom). The degradation model comprises five steps, where ``Comp'' denotes the compression. The text representation describes each degradation operation in a discretized manner, \eg, [\textbf{\textit{medium noise}}] for noise operation. Except for the illustrated aligned prompt-degradation sequence, our pipeline supports more flexible degradation and prompt formats.
    (b) An example to display the dataset.}
    \label{fig-2}
    \vspace{-3.mm}
\end{figure*}

\section{Related Work}
\subsection{Image Super-Resolution} 
\vspace{-1.mm}
Numerous deep networks~\cite{zhang2018image,chen2022cross} have been proposed to advance the field of image SR since the pioneering work of SRCNN~\cite{dong2014learning}. Meanwhile, to enhance the applicability of SR methods in complex (\eg, real-world) applications, blind SR methods have been introduced. To this end, researchers have explored various directions~\cite{liu2022blind}.
First, explicit methods predict the degradation parameters (\eg, blur kernel or noise) as the additional condition for SR networks~\cite{gu2019blind,bell2019blind,zhang2020deep}. For instance, SRMD~\cite{zhang2018learning} takes the LR image with an estimated degradation map for SR reconstruction. 
Second, implicit methods learn underlying degradation models from external datasets~\cite{bulat2018learn}. These methods include supervised learning using paired HR-LR datasets, such as LP-KPN~\cite{cai2019toward}.
Third, simulating real-world degradation with a complex degradation model and synthesize datasets for supervised training~\cite{zhang2023crafting,chen2022real}. For example, Real-ESRGAN~\cite{wang2021real} introduces a high-order degradation model, while BSRGAN~\cite{zhang2021designing} proposes a random shuffling strategy.
However, most methods still face challenges in degradation modeling, thus restricting SR performance.

\subsection{Diffusion Model} 
\vspace{-1.mm}
The diffusion model (DM) has shown significant effectiveness in various synthetic tasks, including image~\cite{ho2020denoising,song2020denoising}, video~\cite{bar2022text2live}, audio~\cite{kong2020diffwave}, and text~\cite{li2022diffusion}. Concurrently, DM has made notable advancements in image manipulation and restoration tasks, such as image editing~\cite{avrahami2022blended}, inpainting~\cite{lugmayr2022repaint}, and deblurring~\cite{whang2022deblurring}. In the field of SR, exploration has also been undertaken. SR3~\cite{saharia2022image} conditions DM with LR images to constrain output space. 
Moreover, some methods~\cite{kawar2022denoising,wang2023zero} apply degradation priors to guide the reverse process of pre-trained DM. However, these methods are primarily tailored for known degradations (\eg, bicubic interpolation). Currently, some approaches~\cite{wang2023exploiting,sun2023coser,yang2024pixel,lin2023diffbir} leverage pre-trained DM and fine-tune it for real SR. Nevertheless, these methods still mainly employ LR images, disregarding the utilization of other modalities (\eg, text) to provide priors.

\subsection{Text Prompt Image Processing}
This field, which includes image generation and image manipulation, is rapidly evolving. For generation, the large-scale text-to-image (T2I) models are successfully constructed using the 
diffusion model and CLIP~\cite{radford2021learning}, \eg, Stable Diffusion~\cite{rombach2022high} and DALL-E-2~\cite{ramesh2022hierarchical}. 
Imagen~\cite{saharia2022photorealistic} further demonstrates the effectiveness of large pre-trained language models, \ie, T5~\cite{raffel2020exploring}, as text encoders. 
Moreover, some methods~\cite{zhang2023adding,qin2023unicontrol}, like ControlNet~\cite{zhang2023adding}, integrate more conditioning controls into text-to-image processes, enabling finer-grained generation.
For manipulation, numerous methods~\cite{hertz2022prompt,kawar2023imagic,kim2022diffusionclip,avrahami2022blended,brooks2023instructpix2pix} have been proposed. For instance, StyleCLIP~\cite{patashnik2021styleclip} combines StyleGAN~\cite{karras2019style} and CLIP~\cite{radford2021learning} to manipulate images using textual descriptions.
DiffusionCLIP~\cite{kim2022diffusionclip} edits global aspects through CLIP gradients~\cite{song2020denoising}. 
Meanwhile, several methods are based on pre-trained T2I models, \eg, Stable Diffusion. For example, Prompt-to-Prompt~\cite{hertz2022prompt} edits synthesis images by modifying text prompts. 
Imagic~\cite{kawar2023imagic} achieves manipulation of real images by fine-tuning models on given images.

\section{Method}
We introduce text prompts into image SR to enhance the reconstruction results.
Our design encompasses two aspects: the dataset and the model.
\textbf{(1) Dataset:} We propose a text-image generation pipeline integrating text prompts into the SR dataset. 
Leveraging the binning method, we apply the text to realize simplified representations of degradation, and combine it with a degradation model to generate data.
\textbf{(2) Model:} We design the PromptSR for image SR conditioned on both text and image. The network is based on the diffusion model and the pre-trained language model.

\subsection{Text-Image Generation Pipeline}
\label{sec:pipeline}
To realize effective training, and enhance model performance, a substantial amount of text-image data is required. 
However, there is a lack of large-scale multi-modal text-image datasets for the SR task. To address this issue, we design the text-image generate pipeline to produce the datasets ($\mathbf{c}$, [$\mathbf{y}$, $\mathbf{x}$]), as illustrated in Fig.~\ref{fig-2}, where $\mathbf{c}$ is the text prompt describing degradation; [$\mathbf{y}$, $\mathbf{x}$] denotes HR and LR images, respectively. 
The pipeline comprises two components: a \textbf{degradation model} that generates HR-LR image pairs and a \textbf{text representation module} that produces text prompts.

\subsubsection{Degradation Model} 
\label{sec:degradation}
We aim to reconstruct HR images from LR images with complex and unknown degradation. 
To encompass the typical degradations while maintaining design simplicity, we develop the degradation model, as depicted in Fig.~\ref{fig-2}\textcolor{iccvblue}{a}. 
Note that while the degradation process in the illustration is applied sequentially, our degradation pipeline supports the more \textbf{flexible} format, \eg, random degradation sequences and the omission of certain components. 

\vspace{0.2em}
\noindent \emph{\textbf{Blur.}}
We employ two kinds of blur: isotropic and anisotropic Gaussian blur. The blur is modeled as a convolution with a blur kernel. The blur kernel is controlled by two parameters: kernel width $\eta$ and standard deviation $\sigma$.

\vspace{0.2em}
\noindent \emph{\textbf{Resize.}}
We upsample/downsample images using two resize with scale factors $\gamma_1$ and $\gamma_2$, respectively. We employ area, bilinear, and bicubic interpolation. The two-step resizing can broaden the degradation range and enhance the generality of the model. We demonstrate it in Sec.~\ref{sec:ablation}.

\vspace{0.2em}
\noindent \emph{\textbf{Noise.}}
We apply Gaussian and Poisson noise, with noise levels controlled by $\varphi_1$ and $\varphi_2$, respectively. Meanwhile, noise is randomly applied in either RGB or gray format.

\vspace{0.2em}
\noindent \emph{\textbf{Compression.}}
We adopt JPEG compression, a widely used compression standard, for image compression. The quality factor $q$ controls the image compression quality.

Given an HR image $\mathbf{y}$, we determine the degradation by randomly selecting the degradation method (\eg, Gaussian or Poisson noise), and sampling all parameters (\eg, noise level $\mu_1$) from the \textbf{uniform distribution}. Through the determined degradation, we can obtain the corresponding LR image $\mathbf{x}$. Compared to other degradation models (\eg, high-order~\cite{wang2021real} or random shuffle~\cite{zhang2021designing}), our degradation model maintains simplicity while covering broad scenarios. 

\subsubsection{Text Prompt} 
\label{sec:text-prompt}
After generating HR-LR image pairs through the degradation model, we further provide descriptions for each pair as text prompts. Consequently, we incorporate text prompts into the dataset.
This process encompasses two key considerations:
\textbf{(1)} The specific content that should be described;
\textbf{(2)} The user-friendly method for generating corresponding descriptions concisely and effectively.
Given the characteristics of image SR, we utilize text to represent degradation. 
Meanwhile, we represent the degradation via a discretization manner based on the binning method~\cite{zhang2023crafting}.

\vspace{0.2em}
\noindent \emph{\textbf{Text prompt for degradation.}}
Typical text prompt image generation and manipulation methods~\cite{ramesh2022hierarchical,ramesh2021zero,avrahami2022blended} apply text prompts to describe the image content. These prompts often revolve semantic-level interpretation and processing of the image content. However, for the image SR task, it is crucial to prioritize fidelity to the original image. Meanwhile, LR images could provide the majority of the low-frequency information~\cite{zhang2018image} and semantic information~\cite{rombach2022high}.

\begin{figure}[t]
\scriptsize
\centering
\begin{tabular}{cc}
\hspace{-4.mm}
\begin{adjustbox}{valign=t}
\begin{tabular}{c}
\includegraphics[width=0.115\textwidth]{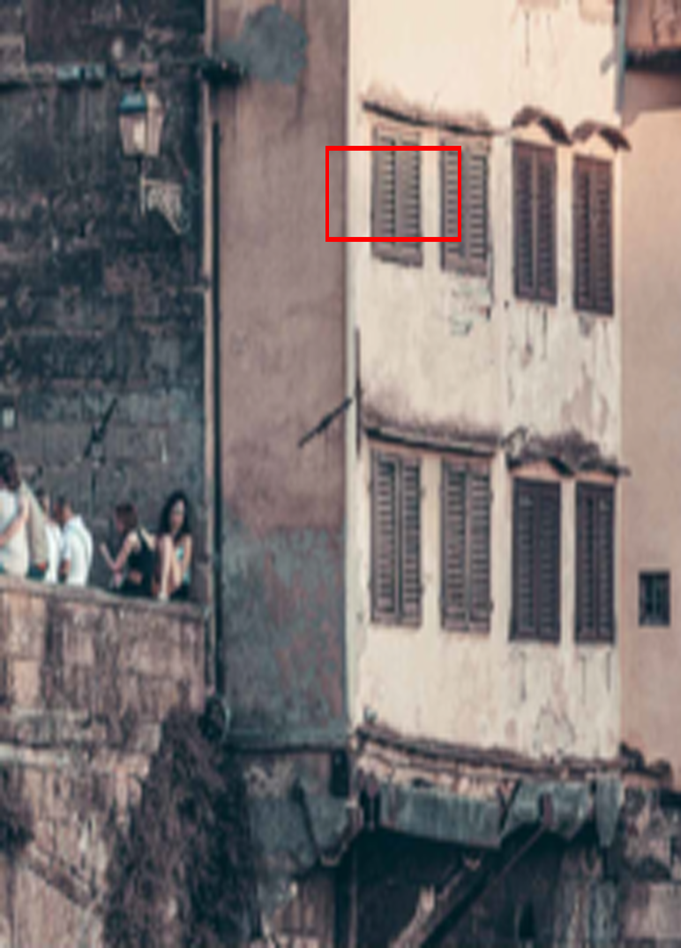}
\\
LR
\end{tabular}
\end{adjustbox}

\hspace{-4.5mm}
\begin{adjustbox}{valign=t}
\begin{tabular}{cccc}
\includegraphics[width=0.174\textwidth]{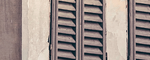} \hspace{-4.mm} &
\includegraphics[width=0.174\textwidth]{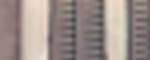} \hspace{-4.mm} &
\\
HR  \hspace{-4.mm} &
Bicubic \hspace{-4.mm} &
\\
\includegraphics[width=0.174\textwidth]{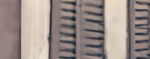} \hspace{-4.mm} &
\includegraphics[width=0.174\textwidth]{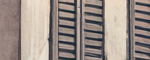} \hspace{-4.mm} &
\\ 
Caption  \hspace{-4.mm} &
Degradation \hspace{-4.mm}
\\
\end{tabular}
\end{adjustbox}

\end{tabular}
\vspace{-3.3mm}
\caption{Visual comparison ($\times$4) of different text contents. Caption (description of the overall image): [\textbf{\textit{people on a weathered balcony of a building with closed shutters}}]. Degradation: [\textbf{\textit{light blur, upsample, light noise, heavy compression, downsample}}].}
\vspace{-5.mm}
\label{fig:prompt}
\end{figure}

\begin{figure*}[t]
    \centering
    \begin{tabular}{c}
    \hspace{-2.mm}\includegraphics[width=1.\linewidth]{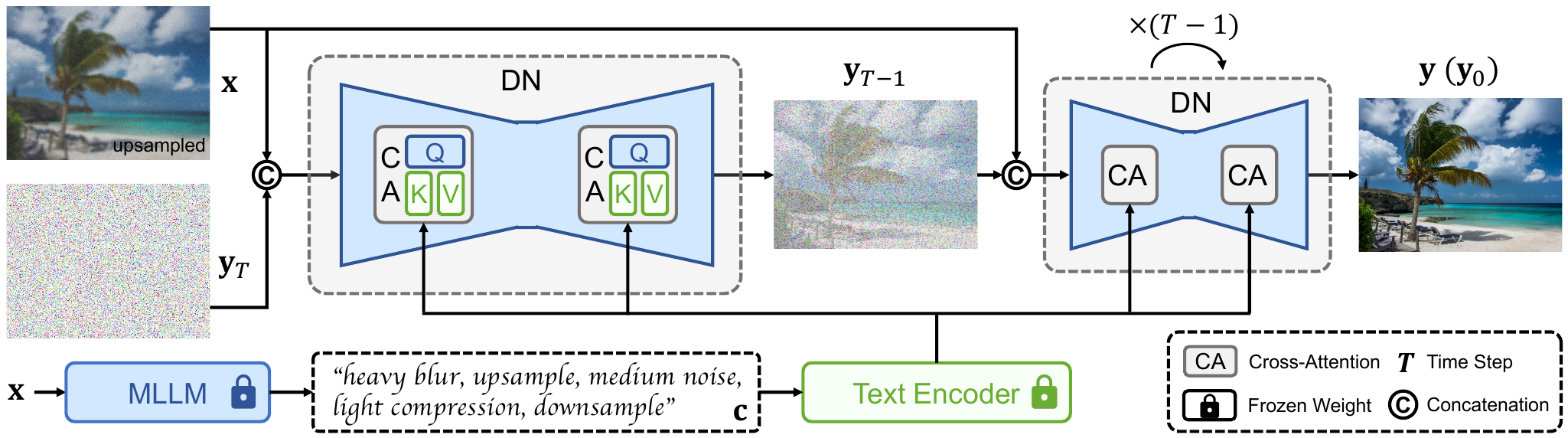} \\
    \end{tabular}
    \vspace{-2.mm}
    \caption{The overall architecture of the PromptSR. The text prompt $\mathbf{c}$ is generated from the LR image $\mathbf{x}$ using the multi-modal large language model (MLLM) and embedded by the pre-trained text encoder. The denoising network (DN) takes the bicubic-\textbf{upsampled} LR image (target HR image resolution), the noise image $\mathbf{y}_t$ ($t\in$$[1, T]$), and the text embedding as inputs to predict noise.}
    \label{fig-3}
    \vspace{-4.mm}
\end{figure*}

Therefore, we adopt the prompt for degradation, instead of the description of the overall image. This prompt can provide degradation priors and thus enhance the capability of methods to model degradation, which is crucial for image SR. 
As shown in Fig.~\ref{fig:prompt}, utilizing text to depict degradation, instead of the overall image content (Caption), yields restoration that is more aligned with the ground truth. To further demonstrate the effectiveness of text prompts for degradation, we provide more analyses in Sec.~\ref{sec:ablation}.

\vspace{0.2em}
\noindent \emph{\textbf{Text representation.}}
To facilitate data generation and practical usability, we describe degradation in natural language with the approach illustrated in Fig.~\ref{fig-2}\textcolor{iccvblue}{a}. Overall, we describe each degradation component via a discretized binning method, and combine them in a flexible format.

First, we discretize the degradation model into several components (\eg, blur) and describe each using qualitative language via a binning method. The sampling distribution of parameters corresponding to each component is evenly divided into discrete intervals (bins). Each bin is summarized to represent the degradation. For instance, we divide the distribution of noise level $\varphi_1$ into three uniform intervals, and describe them as `\textit{light}', `\textit{medium}', and `\textit{heavy}'. Both Gaussian and Poisson noises are summarized as `\textit{noise}', leading to the final representation: [\textbf{\textit{medium noise}}].
Compared to specifying degradation names and their parameters, \eg, [\textbf{\textit{Gaussian noise with noise level 4.5}}], our discretized representation is more \textbf{user-friendly}.

Finally, the overall degradation representation combines all component descriptions, \ie, [\textbf{\textit{deblur description}, ..., \textit{resize description}}]. Figure~\ref{fig-2}\textcolor{iccvblue}{b} illustrates an example. The content of the prompt directly corresponds to the degradation. Furthermore, it is notable that, in our method, the prompt exhibits good generalization and supports \textbf{flexible} description formats. For instance, both arbitrary order or simplified (\eg, only noise description) prompts can still lead to satisfactory restoration outcomes. In Sec.~\ref{sec:ablation}, we conduct a detailed investigation of the prompt format.

\subsection{PromptSR}
PromptSR is based on the general diffusion model~\cite{ho2020denoising}, commonly utilized for high-quality image restoration. The architecture is illustrated in Fig.~\ref{fig-3}.

\subsubsection{Overall Structure}
To underscore the effectiveness of text prompts, we employ a general text-to-image (T2I) diffusion architecture, rather than a meticulously designed structure. Specifically, our method employs a denoising network (DN), operating through a $T$-step reverse process to generate high-resolution (HR) images from Gaussian noise. The DN predicts the noise conditioned on the LR image (upsampled to the target resolution via bicubic interpolation) and text prompt.

For the text prompt, we utilize the multi-modal large language model (MLLM)~\cite{liu2023llava,openai2023gpt4,ye2024mplug} to generate it from the LR image. Concurrently, the pre-trained language model encodes the text prompt. By leveraging the strong capabilities of the language model, our approach achieves a better understanding of image degradation.
For more details on the PromptSR, please refer to the supplementary material.

\subsubsection{MLLM Generation Prompt}
\label{sec:mllm-prompt}
By leveraging the MLLM, users can generate prompts directly from LR images, simplifying the prompt creation process. It also provides a pathway for improving image SR using the MLLM.
Benefiting from the advancements in image quality assessment~\cite{wu2023q,wu2024q,you2024descriptive}, the latest MLLM (\eg, mPLUG-Owl3~\cite{ye2024mplug3}) can generate prompts close to reality. We demonstrate its accuracy in Sec.~\ref{sec:ablation}. Additionally, considering the diverse restoration requirements in actual situations, users can further fine-tune the MLLM-generated prompts to achieve more \textbf{personalized} results. More details are provided in the supplementary material.

\subsubsection{Pre-trained Text Encoder}
Text prompt image models~\cite{patashnik2021styleclip,avrahami2022blended,rombach2022high} mainly employ multi-modal embedding models, \eg, CLIP~\cite{radford2021learning}, as text encoders. These encoders are capable of generating meaningful representations pertinent to tasks. Besides, compared to multi-modal embeddings, pre-trained language models~\cite{devlin2018bert,raffel2020exploring} exhibit stronger text comprehension capabilities. Therefore, we attempt to apply different pre-trained text encoders to build a series of networks. These models demonstrate varying restoration performance levels.

\vspace{-1.mm}
\section{Experiments}
\vspace{-1.mm}
\subsection{Experimental Settings}
\vspace{-1.mm}
\label{sec:setting}
\noindent \textbf{Degradation Settings.}
The degradation model in our proposed pipeline encompasses four operations. Following previous methods~\cite{wang2021real,zhang2021designing}, the parameters for these operations are sampled from the uniform distribution.
\textbf{Blur:} We adopt isotropic Gaussian blur and anisotropic Gaussian blur with equal probability. The kernel width $\eta$ is randomly selected from the set $\{7,9,\dots,21\}$. The standard deviation $\sigma$ is sampled from a uniform distribution $\mathcal{U}_{[0.2, 3]}$.
\textbf{Resize:} We employ area, bilinear, and bicubic interpolation with probabilities of $[0.3,0.4,0.3]$. 
To expand the scope of degradation, we perform two resize operations at different stages. 
The first resize spans upsample and downsample, where the scale factor is $\gamma_1 \sim \mathcal{U}_{[0.15, 1.5]}$. The second resize operation scales the resolution to $\frac{1}{4}$ of the HR image.
\textbf{Noise:} We apply Gaussian and Poisson noise with equal probability. The level of Gaussian noise is $\varphi_1 \sim \mathcal{U}_{[1, 30]}$, while the level of Poisson noise is $\varphi_2 \sim \mathcal{U}_{[0.05, 3]}$.
\textbf{Compression:} We employ JPEG compression with quality factor $q \sim \mathcal{U}_{[30, 95]}$.

\noindent \textbf{Datasets and Metrics.}
We use the LSDIR~\cite{li2023lsdir} as the training dataset. 
We generate the corresponding text-image dataset using our proposed pipeline. We evaluate our method on both synthetic and real-world datasets. For synthetic datasets, we employ Urban100~\cite{huang2015single}, Manga109~\cite{matsui2017sketch}, and the validation (Val) datasets of LSDIR and DIV2K~\cite{timofte2017ntire}. 
For real-world datasets, we utilize RealSR~\cite{cai2019toward}.
We also capture 45 real images directly from the internet, denoted as Real45.
All experiments are conducted on $\times$4 scale.
We adopt two traditional metrics: PSNR and SSIM~\cite{wang2004image} calculated on the Y channel of YCbCr. We also utilize some perceptual metrics: LPIPS~\cite{zhang2018unreasonable}, ST-LPIPS~\cite{ghildyal2022stlpips}, DISTS~\cite{ding2020image}, and CNNIQA~\cite{kang2014convolutional}, and NIMA~\cite{talebi2018nima}. 

\vspace{0.2em}
\noindent \textbf{Implementation Details.}
\label{sec:implementation}
The proposed PromptSR consists of two components: the denoising network (DN) and the pre-trained text encoder. The DN employs a U-Net architecture with a 4-level encoder-decoder. Each level contains two ResNet~\cite{he2016deep,ho2020denoising} blocks and one cross-attention block. 
More details about structure are provided in the supplementary material. 
For the text prompt, we use mPLUG-Owl3~\cite{ye2024mplug3} to generate, and apply the pre-trained multi-modal model, CLIP~\cite{radford2021learning}, to encode. We also discuss other language models as encoders, \eg, T5~\cite{raffel2020exploring}, in Sec.~\ref{sec:ablation}.

We train our model on the generated text-image dataset with a batch size of 16 for a total of 1,000,000 iterations. The input image is randomly cropped to 64$\times$64. We adopt the Adam optimizer~\cite{kingma2014adam} with $\beta_1$=$0.9$ and ${\beta}_2$=$0.99$ to minimize the $\mathcal{L}_1$ loss. The learning rate is 2$\times$10$^{-4}$ and is reduced by half at the 500,000-iteration mark. For DM, we set the total time step $T$ as 2,000. For inference, we employ the DDIM sampling~\cite{song2020denoising} with 50 steps. We use PyTorch~\cite{paszke2019pytorch} to implement our method with 4 Nvidia A100 GPUs.

\begin{table*}[t]
    \hspace{-0.mm}\begin{minipage}{0.53\linewidth}
    \centering
    \resizebox{\columnwidth}{!}{%
    \setlength{\tabcolsep}{4.7mm}
            \begin{tabular}{l | c c c c}
            \toprule
            \rowcolor{color3} & \multicolumn{2}{c}{LSDIR-Val} & \multicolumn{2}{c}{DIV2K-Val} \\
            \rowcolor{color3} \multirow{-2}{*}{\textbf{Method}} & LPIPS~$\textcolor{black}{\downarrow}$ & DISTS~$\textcolor{black}{\downarrow}$ & LPIPS~$\textcolor{black}{\downarrow}$ & DISTS~$\textcolor{black}{\downarrow}$\\
            \midrule
            No Prompt     & 0.3473 & 0.2009 & 0.3384 & 0.1941\\
            MLLM-generated & 0.3261 & 0.1857 & 0.3138 & 0.1750\\
            Pipeline-derived & 0.3211 & 0.1820 & 0.3086 & 0.1727\\
            \bottomrule
            \end{tabular}
    }
    \vspace{-1.mm}
    \caption{Ablation study on the text prompt. We compare an empty string (no prompt) with MLLM-generated and pipeline-derived prompts.}
    \label{tab:ablation-1}
    \end{minipage}
\hfill
    \hspace{-0.mm}\begin{minipage}{0.45\linewidth}
    \centering
    \resizebox{\columnwidth}{!}{
    \setlength{\tabcolsep}{5.mm}
            \begin{tabular}{l | c | c c  c c}
            \toprule
            \rowcolor{color3} \textbf{Method} & \textbf{Metric} & One Resizing & Two Resizings\\
            \midrule
            \multirow{2}{*}{LSDIR-Val} & LPIPS~$\textcolor{black}{\downarrow}$ & 0.3709 & 0.3261  \\
                                      & DISTS~$\textcolor{black}{\downarrow}$ & 0.2254 & 0.1857 \\
            \midrule
            \multirow{2}{*}{DIV2K-Val} & LPIPS~$\textcolor{black}{\downarrow}$ & 0.3570 & 0.3138 \\
            
                                      & DISTS~$\textcolor{black}{\downarrow}$ & 0.2162 & 0.1750 \\
            \bottomrule
            \end{tabular}
    }
    \vspace{-1.mm}
    \caption{Ablation study on the resizing operation. We compare the degradation with one resizing and two resizings.}
    \label{tab:ablation-2-0}
    \end{minipage}
\vspace{-1.mm}
\end{table*}

\begin{figure*}[t]
\scriptsize
\centering
\begin{tabular}{ccc}

\hspace{-4.8mm}
\begin{adjustbox}{valign=t}
\begin{tabular}{cccccc}
\includegraphics[width=0.162\linewidth]{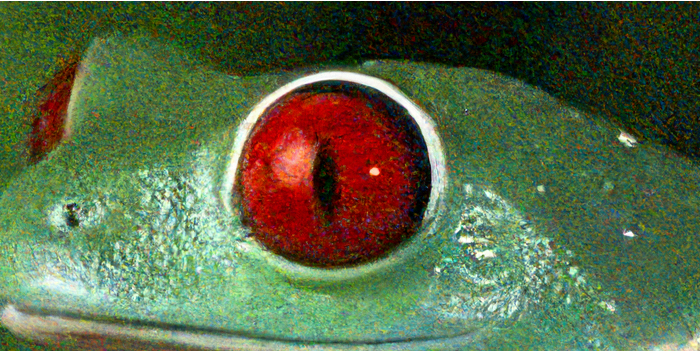} \hspace{-4.2mm} &
\includegraphics[width=0.162\linewidth]{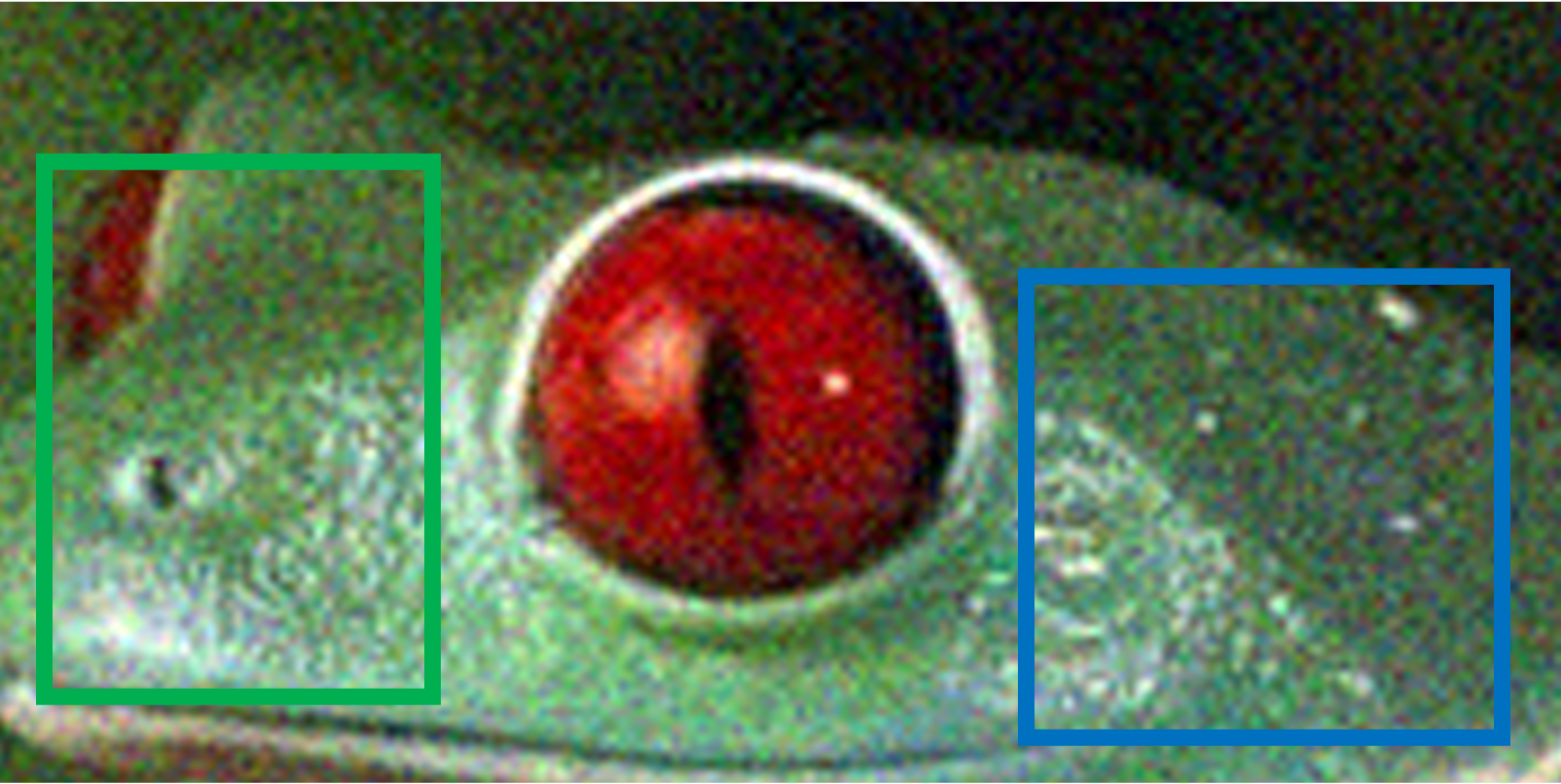} \hspace{-4.2mm} &
\includegraphics[width=0.162\linewidth]{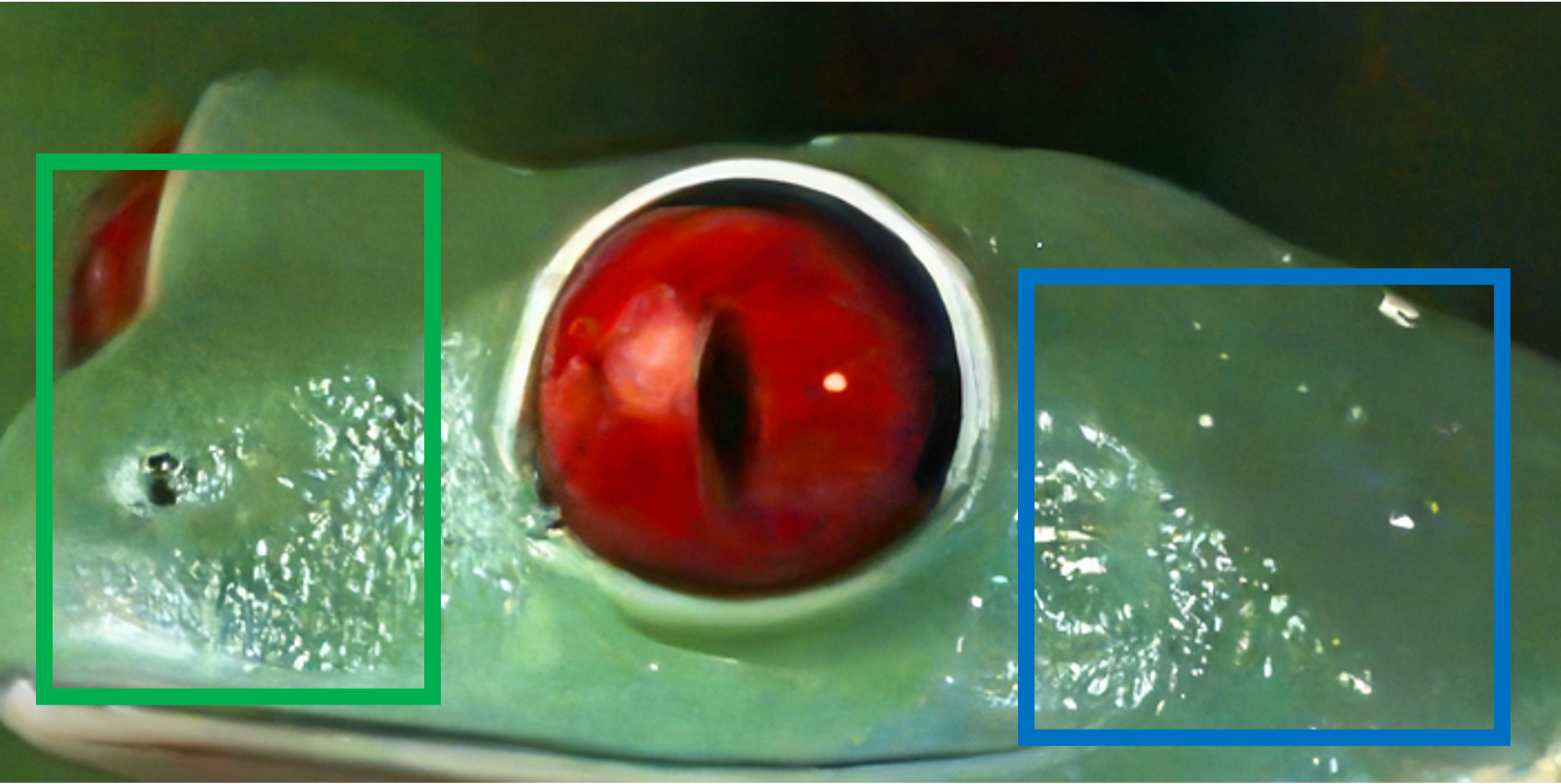} \hspace{-3.8mm} &
\includegraphics[width=0.162\linewidth]{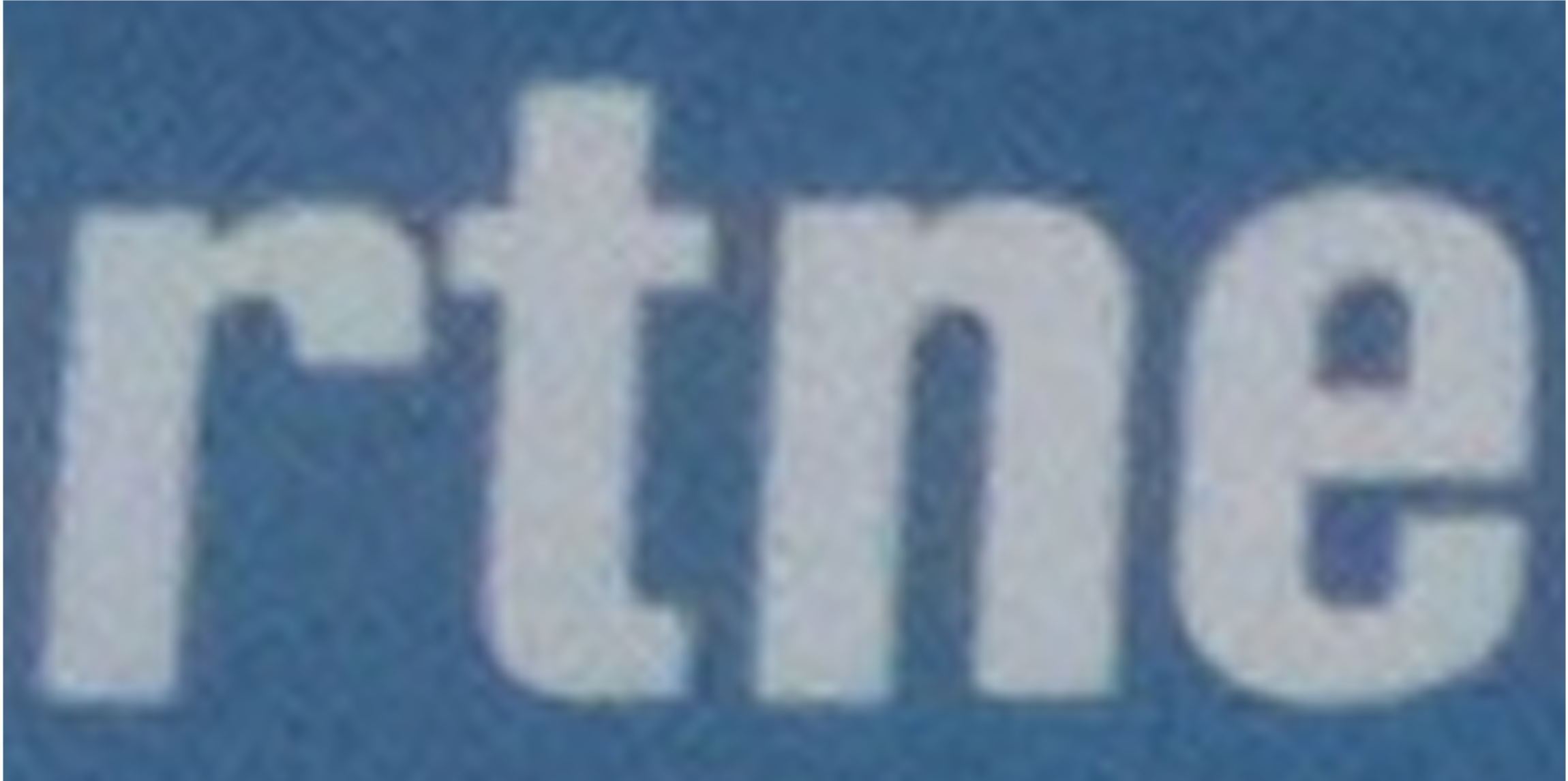} \hspace{-4.2mm} &
\includegraphics[width=0.162\linewidth]{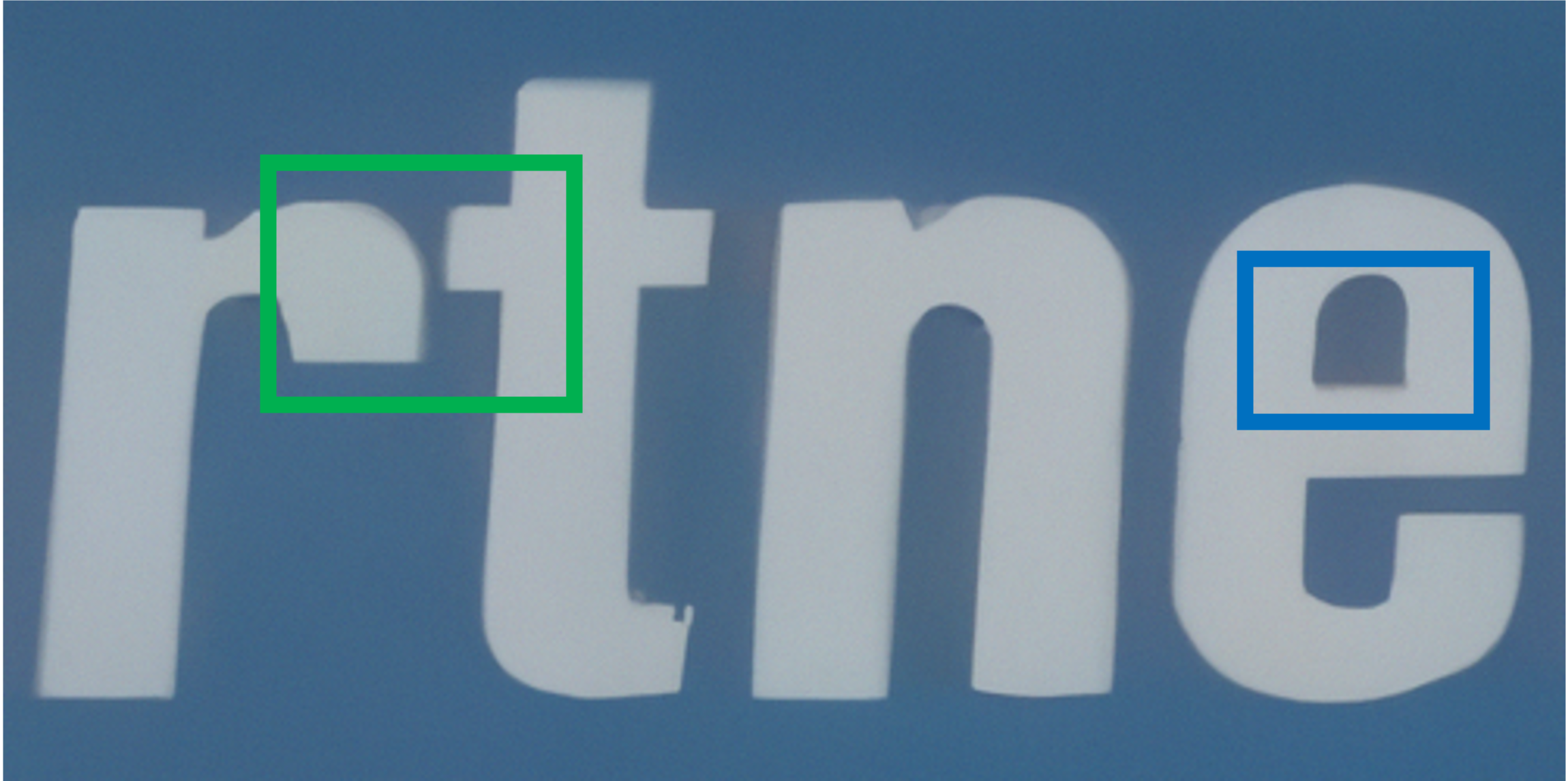} \hspace{-4.2mm} &
\includegraphics[width=0.162\linewidth]{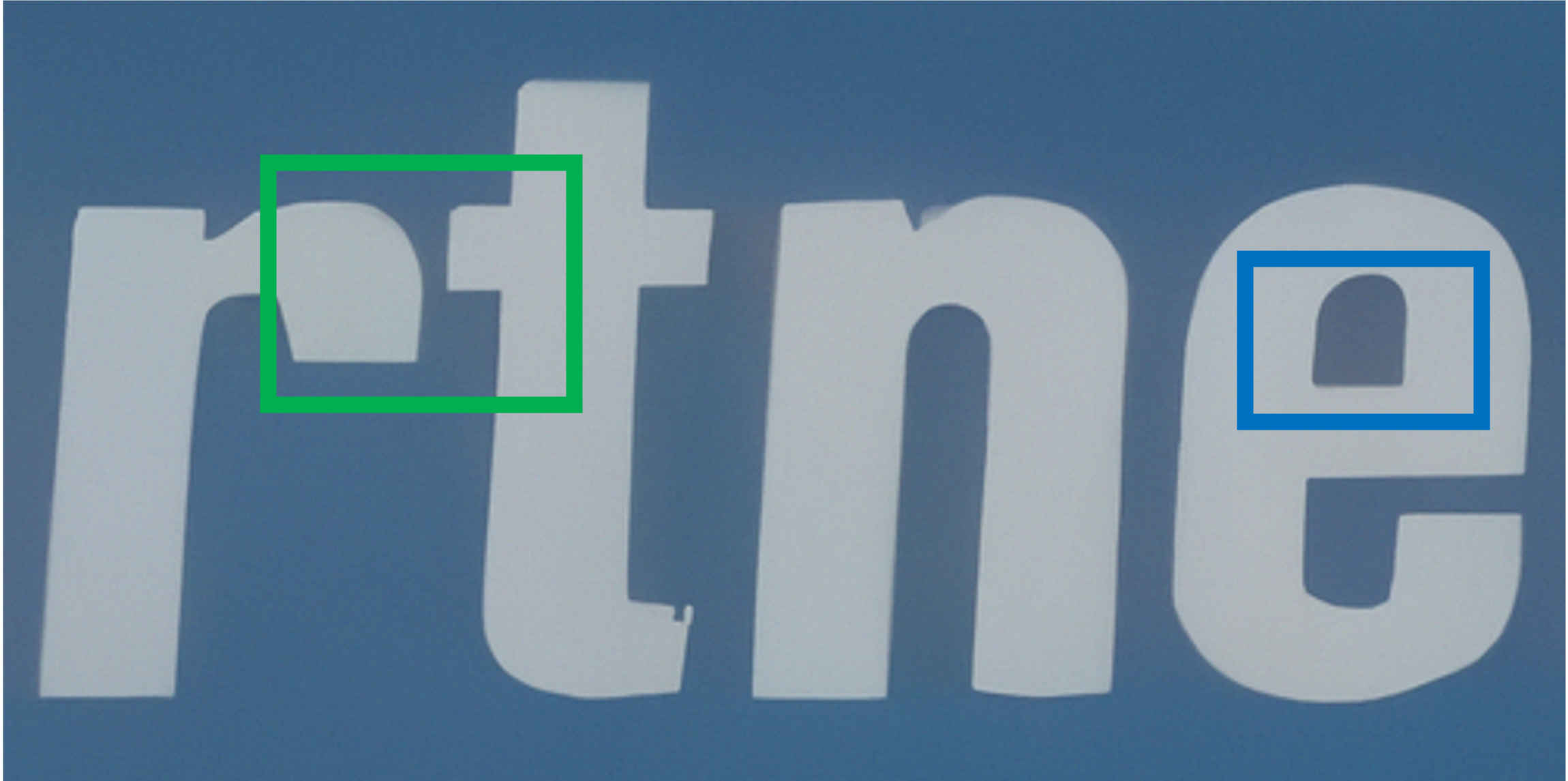} \hspace{-4.2mm} 
\\
LR \hspace{-4.2mm} &
[\textit{light noise}] \hspace{-4.2mm} &
[\textit{heavy noise}] \hspace{-3.9mm} &
LR \hspace{-4.2mm} &
[\textit{medium noise}] \hspace{-4.2mm} &
[\textit{+light blur}] \hspace{-4.2mm}
\\
\end{tabular}
\end{adjustbox}
\end{tabular}
\vspace{-3.mm}
\caption{Visual results ($\times$4) of different prompts. [...] represents the prompt. Boxes in the figures highlight the differences in details.}
\vspace{-2.mm}
\label{fig:ablation-1}
\end{figure*}

\begin{table*}[t]
    \hspace{-0.mm}\begin{minipage}{0.41\linewidth}
    \centering
    \resizebox{\columnwidth}{!}{%
    \setlength{\tabcolsep}{3.mm}
            \begin{tabular}{l | c c  c c}
            \toprule
            \rowcolor{color3} & \multicolumn{2}{c}{Random} & \multicolumn{2}{c}{Fixed} \\
            \rowcolor{color3} \multirow{-2}{*}{\textbf{Method}} & LPIPS~$\textcolor{black}{\downarrow}$ & DISTS~$\textcolor{black}{\downarrow}$ & LPIPS~$\textcolor{black}{\downarrow}$ & DISTS~$\textcolor{black}{\downarrow}$\\
            \midrule
            {LSDIR-Val} & 0.3243 & 0.1860 & 0.3211 & 0.1820\\
            \midrule
            {DIV2K-Val} & 0.3193 & 0.1722 & 0.3086 & 0.1727\\
            \bottomrule
            \end{tabular}
    }
    \subcaption{Different degradation formats.}
    \label{tab:ablation-2-1}
    \end{minipage}
\hfill
    \hspace{-0.mm}\begin{minipage}{0.56\linewidth}
    \centering
    \resizebox{\columnwidth}{!}{%
    \setlength{\tabcolsep}{3.mm}
            \begin{tabular}{l | c c  c c  c c}
            \toprule
            \rowcolor{color3} & \multicolumn{2}{c}{Random} & \multicolumn{2}{c}{Simplified} & \multicolumn{2}{c}{Original} \\
            \rowcolor{color3} \multirow{-2}{*}{\textbf{Method}} & LPIPS~$\textcolor{black}{\downarrow}$ & DISTS~$\textcolor{black}{\downarrow}$ & LPIPS~$\textcolor{black}{\downarrow}$ & DISTS~$\textcolor{black}{\downarrow}$ & LPIPS~$\textcolor{black}{\downarrow}$ & DISTS~$\textcolor{black}{\downarrow}$\\
            \midrule
            {LSDIR-Val} & 0.3231 & 0.1835 & 0.3268 & 0.1871 & 0.3211 & 0.1820\\
            \midrule
            {DIV2K-Val} & 0.3095 & 0.1730 & 0.3131 & 0.1767 & 0.3086 & 0.1727\\
            \bottomrule
            \end{tabular}
    }
    \subcaption{Different prompt formats.}
    \label{tab:ablation-2-2}
    \end{minipage}
    \vspace{-3.mm}
    \caption{Ablation study on the format. 
    (a) Random: shuffled degradation sequence. Fixed: fixed degradation sequence.
    (b) Random: mismatched prompt-degradation order. Simplified: randomly omitting 50\% prompt contents. Original: aligned prompt-degradation order.}
    \label{tab:ablation-2}
\vspace{-1mm}
\end{table*}

\begin{table*}[t]
    \hspace{-0.mm}\begin{minipage}{0.47\linewidth}
    \centering
    \resizebox{\columnwidth}{!}{%
    \setlength{\tabcolsep}{5.0mm}
            \begin{tabular}{l | c c c c}
            \toprule
            \rowcolor{color3} & \multicolumn{2}{c}{LSDIR-Val} & \multicolumn{2}{c}{DIV2K-Val} \\
            \rowcolor{color3} \multirow{-2}{*}{\textbf{Method}} & LPIPS~$\textcolor{black}{\downarrow}$ & DISTS~$\textcolor{black}{\downarrow}$ & LPIPS~$\textcolor{black}{\downarrow}$ & DISTS~$\textcolor{black}{\downarrow}$\\
            \midrule
            Caption     & 0.3403 & 0.1931 & 0.3237 & 0.1840\\
            Degradation & 0.3261 & 0.1857 & 0.3138 & 0.1750\\
            Both        & 0.3247 & 0.1884 & 0.3104 & 0.1770\\
            \bottomrule
            \end{tabular}
    }
    \vspace{-3.mm}
    \caption{Ablation study on the text content. Caption: image content generated by BLIP~\cite{li2022blip}. Degradation (ours): degradation process. Both: the combination of two.}
    \label{tab:ablation-3}
    \end{minipage}
\hfill
    \hspace{-0.mm}\begin{minipage}{0.50\linewidth}
    \centering
    \resizebox{\columnwidth}{!}{
    \setlength{\tabcolsep}{4.mm}
            \begin{tabular}{l | c | c c c c c c}
            \toprule
            \rowcolor{color3} & & \multicolumn{2}{c}{LSDIR-Val} & \multicolumn{2}{c}{DIV2K-Val} \\
            \rowcolor{color3} \multirow{-2}{*}{\textbf{Method}} & \multirow{-2}{*}{\textbf{Params}} & LPIPS~$\textcolor{black}{\downarrow}$ & DISTS~$\textcolor{black}{\downarrow}$ & LPIPS~$\textcolor{black}{\downarrow}$ & DISTS~$\textcolor{black}{\downarrow}$\\
            \midrule
            T5-small & 60M & 0.3260 & 0.1911 & 0.3218 & 0.1863\\
            CLIP     & 428M & 0.3261 & 0.1857 & 0.3138 & 0.1750\\
            T5-xl       & 3B & 0.3151 & 0.1753 & 0.3056 & 0.1682\\
            \bottomrule
            \end{tabular}
    }
    \vspace{-3.mm}
    \caption{Ablation study on the pre-trained text encoder. We adopt different pre-trained language models as text encoders in our PromptSR. Params: the parameters of each text encoder.}
    \label{tab:ablation-4}
    \end{minipage}
\vspace{-1mm}
\end{table*}

\vspace{-1.5mm}
\subsection{Ablation Study}
\vspace{-1.5mm}
\label{sec:ablation}
We investigate the effects of our proposed method at SR ($\times$4) task. We train all models on the LSDIR dataset with 500,000 iterations. We apply the validation datasets of LSDIR~\cite{li2023lsdir} and DIV2K~\cite{timofte2017ntire} for testing. 

\noindent \textbf{Impact of Text Prompt.}
We conduct an ablation to show the influence of introducing the text prompt into image SR. The results are listed in Tab.~\ref{tab:ablation-1}. 
The MLLM-generated prompt is obtained from the LR image as depicted in Sec.~\ref{sec:mllm-prompt}.
The pipeline-derived prompt is directly extracted from the generation pipeline, serving as the ground truth.
The results show that MLLM-generated prompts perform closely to pipeline-derived prompts, and outperform the no-prompt setting. This demonstrates the effectiveness of text prompts and the reliability of the MLLM-generated process.

Moreover, we visualize the impact of different prompts on the SR results in Fig.~\ref{fig:ablation-1}. 
We observe that the method can remove part of the noise for the image with severe noise when the prompt indicates [\textbf{\textit{light noise}}] in the left instance. Conversely, a suitable prompt, \ie, [\textbf{\textit{heavy noise}}], can restore a more realistic result.
Meanwhile, for images at the right, a simplified prompt, \ie, [\textbf{\textit{medium noise}}], can yield a relatively \textbf{satisfactory result}.
Further refining the prompt, \ie, [\textbf{\textit{+light blur}}], can further improve the outcome.
These results indicate the flexibility of our prompts.

\vspace{0.2em}  
\noindent \textbf{Two Resizing Operations.}
We investigate the different number of resizing operations in the degradation. The results are presented in Tab.~\ref{tab:ablation-2-0}. We can find that the model with two resizings performs better.
This is because one single resizing is fixed at $\frac{1}{4}$ in the $\times$4 SR task. Introducing an additional resizing allows for variable scales, expands the degradation scope, and enhances the model generality.

\vspace{0.2em}
\noindent \textbf{Flexible Format.}
We investigate the different formats of the degradation and prompt (pipeline derived) in Tab.~\ref{tab:ablation-2}. Firstly, in Tab.~\ref{tab:ablation-2-1}, we compare fixed and random degradation orders.
The results indicate that random order slightly lowers performance. It may be because random order expands the degradation space (generalization), thus increasing training complexity. 
To balance performance and generalization, we opt for the fixed order shown in Fig.~\ref{fig-2}.

Secondly, in Tab.~\ref{tab:ablation-2-2}, we compare three prompt formats.
The comparison shows that complete prompts (Original) reveal the best performance. Meanwhile, prompt order has little effect. Moreover, the simplified prompt can yield relatively good results due to the model generalization.
Overall, our method exhibits fine generalization, supporting a flexible variety of degradation and prompt.

\vspace{0.2em}
\noindent \textbf{Text Prompt for Degradation.}
We study the effects of different content of text prompts. The results are presented in Tab.~\ref{tab:ablation-3}. We compare three types of text prompt content. 
All experiments are conducted on our proposed PromptSR.
The comparison shows that descriptions of degradation (Degradation) are more suitable for the SR task than image content descriptions (Caption). This is consistent with our analysis in Sec.~\ref{sec:text-prompt}. 
Additionally, combining both descriptions results in a slight performance drop in some metrics compared to using degradation prompts alone. This could be due to the disparity between the two descriptions, which hinders the utilization of degradation information.

\begin{table*}[!t]
\begin{center}
\resizebox{\linewidth}{!}{
\setlength{\tabcolsep}{4.2mm} 
\begin{tabular}{l |c | ccccccc | cc}
        \toprule[0.15em]
        \rowcolor{color3} & & Real-ESRGAN+ & SwinIR-GAN & FeMaSR & Stable Diffusion & StableSR & PASD & DiffBIR & PromptSR\\
        \rowcolor{color3} \multirow{-2}{*}{\textbf{Dataset}} & \multirow{-2}{*}{\textbf{Metric}}  & \cite{wang2021real} & \cite{liang2021swinir} & \cite{chen2022real} & \cite{rombach2022high} & \cite{wang2023exploiting} & \cite{yang2024pixel} & \cite{lin2023diffbir} & (ours)\\
        \midrule
        \multirow{7}{*}{Urban100} & PSNR~$\textcolor{black}{\uparrow}$ & 20.89 & 20.91 & 20.37 & 20.20 & 18.25 & 21.32 & \textcolor{red}{21.73} & \textcolor{blue}{21.39}\\
                  & SSIM~$\textcolor{black}{\uparrow}$ & 0.5997 & \textcolor{blue}{0.6013} & 0.5573 & 0.4852 & 0.4454 & 0.5897 & 0.5896 & \textcolor{red}{0.6130}\\
                  & LPIPS~$\textcolor{black}{\downarrow}$ & 0.2621 & {0.2547} & 0.2725 & 0.4589 & 0.3941 & \textcolor{blue}{0.2522} & 0.2586 & \textcolor{red}{0.2500}\\
                  & ST-LPIPS~$\textcolor{black}{\downarrow}$ & 0.2494 & \textcolor{blue}{0.2376} & {0.2442} & 0.3845 & 0.3301 & 0.2501 & 0.2686 & \textcolor{red}{0.2262}\\
                  & DISTS~$\textcolor{black}{\downarrow}$ & \textcolor{blue}{0.1762} & \textcolor{red}{0.1676} & 0.1877 & 0.2505 & 0.2153 & 0.1952 & {0.1857} & {0.1857}\\
                  & CNNIQA~$\textcolor{black}{\uparrow}$ & 0.6635 & 0.6614 & \textcolor{red}{0.6781} & 0.5870 & 0.4684 & 0.6560 & 0.6517 & \textcolor{blue}{0.6732}\\
                  & NIMA~$\textcolor{black}{\uparrow}$ & 5.3135 & 5.3622 & {5.4161} & 4.6368 & \textcolor{blue}{5.5172} & \textcolor{red}{5.8362} & 5.4010 & {5.5059}\\
        \midrule
        \multirow{7}{*}{Manga109} & PSNR~$\textcolor{black}{\uparrow}$ & \textcolor{blue}{21.62} & \textcolor{red}{21.81} & 21.46 & 18.76 & 14.23 & 21.30 & 21.37 & 20.82\\
                  & SSIM~$\textcolor{black}{\uparrow}$ & \textcolor{blue}{0.7217} & \textcolor{red}{0.7258} & 0.6891 & 0.5412 & 0.4120 & 0.6960 & 0.6738 & 0.7048\\
                  & LPIPS~$\textcolor{black}{\downarrow}$ & 0.2051 & \textcolor{blue}{0.2047} & 0.2145 & 0.3699 & 0.4670 & 0.2180 & 0.2198 & \textcolor{red}{0.1856}\\
                  & ST-LPIPS~$\textcolor{black}{\downarrow}$ & 0.1649 & 0.1590 & \textcolor{blue}{0.1520} & 0.2750 & 0.3568 & 0.1694 & 0.1679 & \textcolor{red}{0.1205}\\
                  & DISTS~$\textcolor{black}{\downarrow}$ & \textcolor{blue}{0.1252} & \textcolor{red}{0.1185} & 0.1418 & 0.1638 & 0.1784 & 0.1389 & 0.1380 & 0.1373\\
                  & CNNIQA~$\textcolor{black}{\uparrow}$ & 0.6651 & 0.6673 & 0.6735 & 0.6691 & 0.4048 & 0.6262 & \textcolor{red}{0.6988} & \textcolor{blue}{0.6929}\\
                  & NIMA~$\textcolor{black}{\uparrow}$ & 4.9825 & 4.8784 & 5.0625 & 4.6493 & {4.7099} & 5.1632 & \textcolor{blue}{5.1738} & \textcolor{red}{5.4211}\\
        \midrule
        \multirow{7}{*}{LSDIR-Val} & PSNR~$\textcolor{black}{\uparrow}$ & 22.40 & 22.34 & 21.19 & 19.91 & 16.83 & \textcolor{blue}{22.45} & \textcolor{red}{22.63} & 22.44\\
                  & SSIM~$\textcolor{black}{\uparrow}$ & \textcolor{red}{0.6115} & 0.6067 & 0.5542 & 0.4487 & 0.3635 & 0.5851 & 0.5725 & \textcolor{blue}{0.6070}\\
                  & LPIPS~$\textcolor{black}{\downarrow}$ & 0.2932 & \textcolor{blue}{0.2911} & 0.2917 & 0.4489 & 0.5167 & 0.3089 & 0.3104 & \textcolor{red}{0.2810}\\
                  & ST-LPIPS~$\textcolor{black}{\downarrow}$ & 0.2502 & {0.2440} & \textcolor{blue}{0.2362} & 0.3521 & 0.3679 & 0.2591 & 0.2827 & \textcolor{red}{0.2258}\\
                  & DISTS~$\textcolor{black}{\downarrow}$ & 0.1627 & 0.1598 & \textcolor{red}{0.1533} & 0.2240 & 0.2248 & 0.1698 & 0.1758 & \textcolor{blue}{0.1548}\\
                  & CNNIQA~$\textcolor{black}{\uparrow}$ & 0.6417 & 0.6277 & \textcolor{blue}{0.6716} & 0.6563 & 0.3991 & 0.6314 & 0.5339 & \textcolor{red}{0.6726}\\
                  & NIMA~$\textcolor{black}{\uparrow}$ & 4.9878 & 4.9551 & {5.1998} & 4.4452 & 5.1989 & \textcolor{red}{5.7027} & 5.1883 & \textcolor{blue}{5.2538}\\
        \midrule
        \multirow{7}{*}{DIV2K-Val} & PSNR~$\textcolor{black}{\uparrow}$ & 25.24 & \textcolor{red}{25.73} & 23.80 & 21.47 & 17.53 & 25.16 & \textcolor{blue}{25.56} & 25.14\\
                  & SSIM~$\textcolor{black}{\uparrow}$ & \textcolor{red}{0.7017} & \textcolor{blue}{0.6932} & 0.6310 & 0.5120 & 0.4034 & 0.6747 & 0.6653 & 0.6813\\
                  & LPIPS~$\textcolor{black}{\downarrow}$ & 0.2896 & \textcolor{blue}{0.2854} & 0.2899 & 0.4709 & 0.5617 & 0.3042 & 0.2973 & \textcolor{red}{0.2753}\\
                  & ST-LPIPS~$\textcolor{black}{\downarrow}$ & 0.2186 & 0.2090 & \textcolor{blue}{0.2061} & 0.2307 & 0.4098 & 0.2191 & 0.3717 & \textcolor{red}{0.1913}\\
                  & DISTS~$\textcolor{black}{\downarrow}$ & 0.1548 & 0.1497 & \textcolor{red}{0.1451} & 0.2239 & 0.2316 & 0.1620 & 0.1809 & \textcolor{blue}{0.1484}\\
                  & CNNIQA~$\textcolor{black}{\uparrow}$ & 0.6238 & 0.6125 & \textcolor{blue}{0.6617} & 0.5814 & 0.3799 & 0.6357 & 0.6380 & \textcolor{red}{0.6748}\\
                  & NIMA~$\textcolor{black}{\uparrow}$ & 4.8202 & 4.8015 & {5.0451} & 4.3881 & \textcolor{blue}{5.2690} & \textcolor{red}{5.6535} & 5.0213 & {5.0834}\\
        \bottomrule[0.15em]
        
\end{tabular}
}
\vspace{-2.mm}
\caption{Quantitative comparison ($\times$4) on synthetic datasets. The best and second-best results are colored \textcolor{red}{red} and \textcolor{blue}{blue}.}
\label{tab:compare-1}
\end{center}
\vspace{-9.mm}
\end{table*}

\vspace{0.2em}
\noindent \textbf{Pre-trained Text Encoder.}
We further explore the impact of different text encoders, with the results detailed in Tab.~\ref{tab:ablation-4}. We utilize several pre-trained text encoders: \textbf{CLIP}~\cite{radford2021learning} (clip-vit-large) and \textbf{T5}~\cite{raffel2020exploring} (T5-small and T5-xl).
We discover that models employing different text encoders display varied performance. Applying more powerful language models as text encoders enhances model performance. For instance, T5-xl, compared to T5-small, reduces the LPIPS on the LSDIR and DIV2K validation sets by 0.0109 and 0.0162, respectively. Moreover, it is also notable that the performance of the model is not entirely proportional to the parameter size of the text encoder. Considering both model performance and parameter size, we apply CLIP.

\subsection{Evaluation on Synthetic Datasets}
\begin{figure*}[!t]
\scriptsize
\centering
\begin{tabular}{cccccccc}

\hspace{-3.mm}
\begin{adjustbox}{valign=t}
\begin{tabular}{c}
\includegraphics[width=0.217\textwidth]{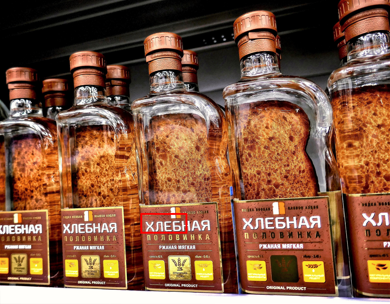}
\\
LSDIR-Val
\end{tabular}
\end{adjustbox}
\hspace{-4.5mm}
\begin{adjustbox}{valign=t}
\begin{tabular}{cccccc}
\includegraphics[width=0.185\textwidth]{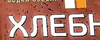} \hspace{-4.mm} &
\includegraphics[width=0.185\textwidth]{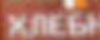} \hspace{-4.mm} &
\includegraphics[width=0.185\textwidth]{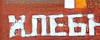} \hspace{-4.mm} &
\includegraphics[width=0.185\textwidth]{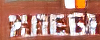} \hspace{-4.mm} &
\\ 
HR \hspace{-4mm} &
Bicubic \hspace{-4mm} &
SwinIR-GAN~\cite{liang2021swinir} \hspace{-4mm} &
FeMaSR~\cite{chen2022real} \hspace{-4mm} &
\\
\includegraphics[width=0.185\textwidth]{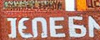} \hspace{-4.mm} &
\includegraphics[width=0.185\textwidth]{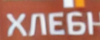} \hspace{-4.mm} &
\includegraphics[width=0.185\textwidth]{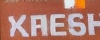} \hspace{-4.mm} &
\includegraphics[width=0.185\textwidth]{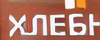} \hspace{-4.mm} &
\\ 
Stable Diffusion~\cite{rombach2022high} \hspace{-4mm} &
PASD~\cite{yang2024pixel} \hspace{-4mm} &
DiffBIR~\cite{lin2023diffbir} \hspace{-4mm} &
PromptSR (ours) \hspace{-4mm}
\\
\end{tabular}
\end{adjustbox}
\\

\hspace{-3.mm}
\begin{adjustbox}{valign=t}
\begin{tabular}{c}
\includegraphics[width=0.217\textwidth]{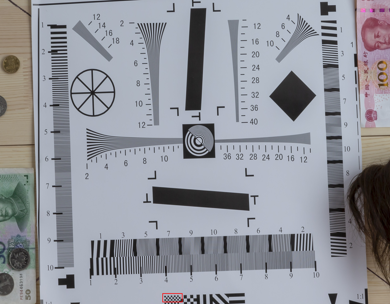}
\\
RealSR
\end{tabular}
\end{adjustbox}
\hspace{-4.5mm}
\begin{adjustbox}{valign=t}
\begin{tabular}{cccccc}
\includegraphics[width=0.185\textwidth]{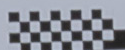} \hspace{-4.mm} &
\includegraphics[width=0.185\textwidth]{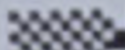} \hspace{-4.mm} &
\includegraphics[width=0.185\textwidth]{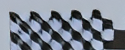} \hspace{-4.mm} &
\includegraphics[width=0.185\textwidth]{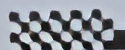} \hspace{-4.mm} &
\\ 
HR \hspace{-4mm} &
Bicubic \hspace{-4mm} &
SwinIR-GAN~\cite{liang2021swinir} \hspace{-4mm} &
FeMaSR~\cite{chen2022real} \hspace{-4mm} &
\\
\includegraphics[width=0.185\textwidth]{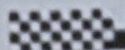} \hspace{-4.mm} &
\includegraphics[width=0.185\textwidth]{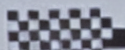} \hspace{-4.mm} &
\includegraphics[width=0.185\textwidth]{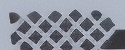} \hspace{-4.mm} &
\includegraphics[width=0.185\textwidth]{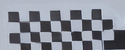} \hspace{-4.mm} &
\\ 
Stable Diffusion~\cite{rombach2022high} \hspace{-4mm} &
PASD~\cite{yang2024pixel} \hspace{-4mm} &
DiffBIR~\cite{lin2023diffbir} \hspace{-4mm} &
PromptSR (ours) \hspace{-4mm}
\\
\end{tabular}
\end{adjustbox}
\\

\hspace{-3.mm}
\begin{adjustbox}{valign=t}
\begin{tabular}{c}
\includegraphics[width=0.217\textwidth]{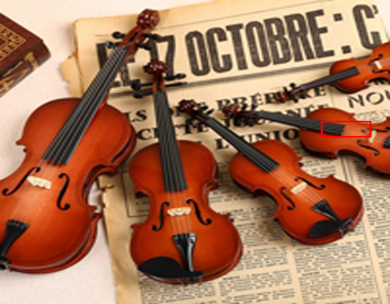}
\\
Real45
\end{tabular}
\end{adjustbox}
\hspace{-4.5mm}
\begin{adjustbox}{valign=t}
\begin{tabular}{cccccc}
\includegraphics[width=0.185\textwidth]{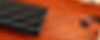} \hspace{-4.mm} &
\includegraphics[width=0.185\textwidth]{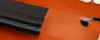} \hspace{-4.mm} &
\includegraphics[width=0.185\textwidth]{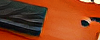} \hspace{-4.mm} &
\includegraphics[width=0.185\textwidth]{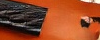} \hspace{-4.mm} &
\\ 
LR \hspace{-4mm} &
Real-ESRGAN+~\cite{wang2021real} \hspace{-4mm} &
SwinIR-GAN~\cite{liang2021swinir} \hspace{-4mm} &
FeMaSR~\cite{chen2022real} \hspace{-4mm} &
\\
\includegraphics[width=0.185\textwidth]{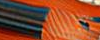} \hspace{-4.mm} &
\includegraphics[width=0.185\textwidth]{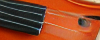} \hspace{-4.mm} &
\includegraphics[width=0.185\textwidth]{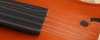} \hspace{-4.mm} &
\includegraphics[width=0.185\textwidth]{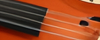} \hspace{-4.mm} &
\\ 
Stable Diffusion~\cite{rombach2022high} \hspace{-4mm} &
PASD~\cite{yang2024pixel} \hspace{-4mm} &
DiffBIR~\cite{lin2023diffbir} \hspace{-4mm} &
PromptSR (ours) \hspace{-4mm}
\end{tabular}
\end{adjustbox}

\end{tabular}
\vspace{-2.mm}
\caption{Visual comparison ($\times$4) on synthetic and real-world datasets. Our method can generate more realistic images.}
\label{fig:visual}
\vspace{-4.mm}
\end{figure*}

We compare our proposed method with several recent state-of-the-art methods: Real-ESRGAN+~\cite{wang2021real}, SwinIR-GAN~\cite{liang2021swinir}, FeMaSR~\cite{chen2022real}, Stable Diffusion~\cite{rombach2022high}, StableSR~\cite{wang2023exploiting}, PASD~\cite{yang2024pixel}, and DiffBIR~\cite{lin2023diffbir}. 
We show quantitative results in Tab.~\ref{tab:compare-1} and visual results in Fig.~\ref{fig:visual}.

Meanwhile, the text prompt for our method is generated from MLLM as depicted in Sec.~\ref{sec:mllm-prompt}. For instance, the prompt for the first case in Fig.~\ref{fig:visual}: [\textbf{\textit{medium blur, unchange, light noise, heavy compression, downsample}}].

\noindent \textbf{Quantitative Results.}
We evaluate our method on some synthetic test datasets: Urban100~\cite{huang2015single}, Manga109~\cite{matsui2017sketch}, LSDIR-Val~\cite{li2023lsdir}, and DIV2K-Val~\cite{timofte2017ntire} in Tab.~\ref{tab:compare-1}. 
Our method outperforms others on most \textbf{perceptual metrics}.
For instance, compared with DiffBIR~\cite{lin2023diffbir}, our PromptSR achieves a reduction in LPIPS by 0.0294 and 0.0220 on LSDIR-Val and DIV2K-Val, respectively.
Moreover, for PSNR and SSIM, the two metrics are only used as references, since they do not consistently align well with the image quality~\cite{saharia2022image}.
These results demonstrate that introducing text prompts into SR can effectively improve performance.

\vspace{0.2em}
\noindent \textbf{Visual Results.}
We show one visual comparison in the first instance in Fig.~\ref{fig:visual}. 
We can observe that our proposed PromptSR is capable of restoring clearer and more realistic images, in some challenging cases. 
For instance, in the first example, other models (\eg, DiffBIR~\cite{lin2023diffbir}) tend to construct unclear or incorrect text. On the contrary, our approach can reconstruct faithful and realistic results.
This is consistent with the quantitative results. Furthermore, we provide more visual results in the supplementary material.

\begin{table*}[!t]
\small
\begin{center}
\resizebox{\linewidth}{!}{
\setlength{\tabcolsep}{3.5mm} 
\begin{tabular}{l |c | ccccccc | cc}
        \toprule[0.15em]
        \rowcolor{color3} & & Real-ESRGAN+ & SwinIR-GAN & FeMaSR & Stable Diffusion & StableSR & PASD & DiffBIR & PromptSR\\
        \rowcolor{color3} \multirow{-2}{*}{\textbf{Dataset}} & \multirow{-2}{*}{\textbf{Metric}}  & \cite{wang2021real} & \cite{liang2021swinir} & \cite{chen2022real} & \cite{rombach2022high} & \cite{wang2023exploiting} & \cite{yang2024pixel} & \cite{lin2023diffbir} & (ours)\\
        \midrule
        \multirow{7}{*}{RealSR} & PSNR~$\textcolor{black}{\uparrow}$ & 25.62 & 26.54 & 25.74 & 24.11 & 25.34 & \textcolor{blue}{26.96} & \textcolor{red}{27.42} & 26.71\\
                   & SSIM~$\textcolor{black}{\uparrow}$ & 0.7582 & \textcolor{red}{0.7918} & 0.7643 & 0.6980 & 0.7385 & 0.7799 & 0.7790 & \textcolor{blue}{0.7821}\\
                   & LPIPS~$\textcolor{black}{\downarrow}$ & 0.2843 & {0.2765} & 0.2938 & 0.5035 & 0.2961 & \textcolor{blue}{0.2733} & 0.3434 & \textcolor{red}{0.2702}\\
                   & ST-LPIPS~$\textcolor{black}{\downarrow}$ & 0.2165 & 0.2078 & \textcolor{blue}{0.1990} & 0.4122 & 0.2278 & 0.2217 & 0.2506 & \textcolor{red}{0.1937}\\
                   & DISTS~$\textcolor{black}{\downarrow}$ & 0.1732 & \textcolor{red}{0.1672} & 0.1927 & 0.2441 & 0.1904 & 0.1874 & 0.2140 & \textcolor{blue}{0.1820}\\
                   & CNNIQA~$\textcolor{black}{\uparrow}$ & 0.5755 & 0.5208 & \textcolor{blue}{0.5916} & 0.4465 & 0.3227 & 0.5254 & 0.5544 & \textcolor{red}{0.6376}\\
                   & NIMA~$\textcolor{black}{\uparrow}$ & 4.7673 & 4.7338 & 4.8745 & 4.1598 & 4.7933 & \textcolor{red}{5.1036} & 4.8295 & \textcolor{blue}{4.8917}\\
        \bottomrule[0.15em]
        
\end{tabular}
}
\vspace{-2.mm}
\caption{Quantitative comparison ($\times$4) on the real-world dataset. The best and second-best results are colored \textcolor{red}{red} and \textcolor{blue}{blue}.}
\label{tab:compare-2}
\end{center}
\vspace{-5.mm}
\end{table*}

\subsection{Evaluation on Real-World Datasets}
We further evaluate our method on real-world datasets. The prompts also generated by MLLM. Some prompt instances are provided in the supplementary material. The quantitative and visual results are shown in Tab.~\ref{tab:compare-2} and Fig.~\ref{fig:visual}.

\noindent \textbf{Quantitative Results.}
We present the quantitative comparison on RealSR~\cite{cai2019toward} in Tab.~\ref{tab:compare-2}. Our PromptSR achieves the best performance on most perceptual metrics, including LPIPS, ST-LPIPS, and CNNIQA. It also scores well on DISTS and NIMA.
Compared with DiffBIR~\cite{lin2023diffbir}, our method obtains a 0.0732 improvement in LPIPS and a 0.032 gain in DISTS.
These results further demonstrate the superiority of introducing text prompts into image SR tasks.

\vspace{0.2em}
\noindent \textbf{Visual Results.}
We present some visual results in Fig.~\ref{fig:visual}. Except for the RealSR dataset, we also conduct an evaluation on the Real45 dataset, collected from the internet. Our proposed method also outperforms other single-modal methods on real-world datasets. For example, in the second instance, our method can generate clear patterns. However, other methods produce blurred or incorrect (\eg, PASD~\cite{yang2024pixel}) results, or they introduce artifacts (\eg, FeMaSR~\cite{chen2022real}). In the third example, compared methods result in excessive blurring or the presence of artifacts. In contrast, our PromptSR successfully restores the four guitar strings clearly, which is consistent with reality. More comparison results are provided in the supplementary material.

\vspace{-1.mm}
\section{Conclusion}
\vspace{-1.mm}
In this work, we introduce the text prompts to provide degradation priors for enhancing image SR. 
Specifically, we develop a text-image generation pipeline to integrate text into the SR dataset, via text degradation representation and degradation model. The text representation is flexible and user-friendly.
Meanwhile, we propose the PromptSR to realize the text prompt SR. The PromptSR applies the MLLM to generate the degradation prompt, and utilizes the pre-trained language model to enhance text guidance. 
We train our PromptSR on the generated text-image dataset and evaluate it on both synthetic and real-world datasets. Superior quantitative and visual results demonstrate the effectiveness of introducing text prompts into image SR.

{
    \small
    \bibliographystyle{ieeenat_fullname}
    \bibliography{main}
}

\end{document}